# Script Identification in Natural Scene Image and Video Frame using Attention based Convolutional-LSTM Network

[a]Ankan Kumar Bhunia[1], [b]Aishik Konwer[1], [c]Ayan Kumar Bhunia,[d]Abir Bhowmick,[e]Partha P. Roy*, [f]Umapada Pal

[a]Dept. of EE, Jadavpur University, Kolkata, India. Email-[a]ankankumarbhunia@gmail.com
[b]Dept. of ECE, Institute of Engineering & Management, Kolkata, India. Email-[b]konweraishik@gmail.com
[c]Dept. of ECE, Institute of Engineering & Management, Kolkata, India.Email-[c]ayanbhunia007@gmail.com
[d]Dept. of ECE, Institute of Engineering & Management, Kolkata, India. Email-[d]bhowmick.abir@rediffmail.com
[e]Dept. of CSE, Indian Institute of Technology, Roorkee, India. Email-[e]proy.fcs@iitr.ac.in
[f]Umapada Pal, CVPR Unit, Indian Statistical Institute, Kolkata, India. email-[f]umapada@isical.ac.in
[e]TEL: +91-1332-284816

## Abstract

Script identification plays a significant role in analysing documents and videos. In this paper, we focus on the problem of script identification in scene text images and video scripts. Because of low image quality, complex background and similar layout of characters shared by some scripts like Greek, Latin, etc., text recognition in those cases become challenging. In this paper, we propose a novel method that involves extraction of local and global features using CNN-LSTM framework and weighting them dynamically for script identification. First, we convert the images into patches and feed them into a CNN-LSTM framework. Attention-based patch weights are calculated applying softmax layer after LSTM. Next, we do patch-wise multiplication of these weights with corresponding CNN to yield local features. Global features are also extracted from last cell state of LSTM. We employ a fusion technique which dynamically weights the local and global features for an individual patch. Experiments have been done in four public script identification datasets: SIW-13, CVSI2015, ICDAR-17 and MLe2e. The proposed framework achieves superior results in comparison to conventional methods.

## 1. Introduction

Script identification is one of the essential elements of Optical Character Recognition (OCR).

[1] Both the authors contributed equally.

Provided an input text image, the function of script identification is to classify it into one of the available scripts which include English, Chinese, Greek, Arabic etc. Some examples of scene text images from various scripts are shown in Figure 1. Script identification task can be posed as an image classification problem that has been thoroughly studied recently. It is potentially applied for different purposes such as scene understanding [1], image searching of any product [2], mobile phone navigation, video caption recognition [3], and machine translation [4,5].

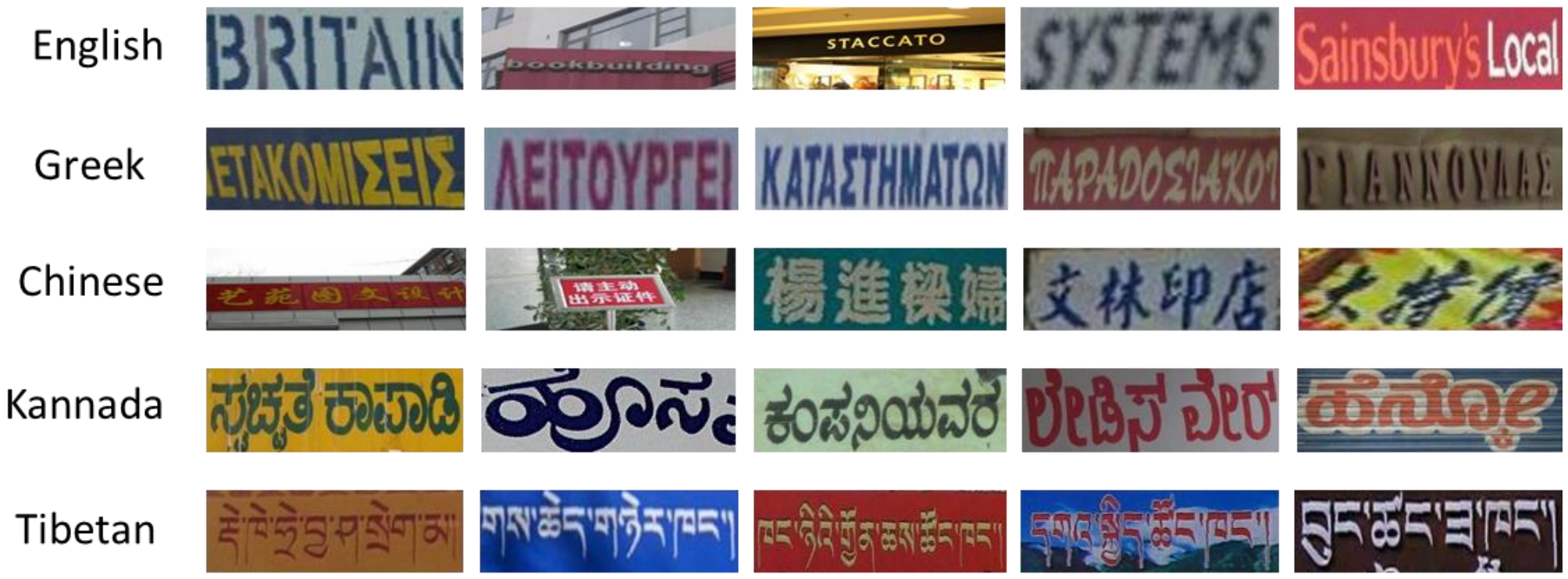

**Fig.1. Examples of scene text images**

In the field of document image analysis problems, script identification has gained plenty of popularity in recent years. Recently, advanced problems like font-to-font translation, staff line removal, handwriting trajectory recovery etc. have been addressed employing deep learning techniques in [66-68]. However the main area of research lies in script identification of printed documents or videos. Spitz in [6] exploits different spatial relationships of features connected to concave shapes in character structures, for page-wise script identification. In [7] the authors addressed text level script identification of Indian language using projection profile. Hochberg et al. [8] developed a novel method utilizing templates based on clusters to deal with distinct characteristic layouts. Tan in [9] employed texture level features unaffected by rotation, for identification of Chinese, English, Greek, Russian and other such text. In [10], Singh et al. make use of mid-level feature representation extracted from densely calculated local features and in end a readymade classifier for script identification from text image. All the above methods have achieved great results but only in document script identification.

However, identification of script from natural scenes is still a thought-provoking problem and has not been dealt with much. As texts in natural scenes often hold productive, high quality information, many works are found in localization and recognition of scene text [11-17]. Script identification in the wild is an unavoidable pre-processing of a multi-lingual scene text understanding scheme [18-20]. But this scene text identification is difficult because its characteristics are quite dissimilar to normal image classification, or document/video script identification, largely owing to the following reasons: Firstly, in natural scenes, text presents more diversity compared to documents or videos. They are frequently spotted on complex backgrounds such as outdoor sign-boards and hoardings, written in different fonts and styles. Fonts and colour of the text have large variations. Secondly, the image quality is often degraded by distortions such as low resolutions, noises, and varying light conditions. This results in low accuracy. Traditional document analysing methods like binarization and component analysis appear untrustworthy. And finally, few languages contain relatively small dissimilarities, e.g. scripts like Greek, English and Russian share a subset of characters that have nearly the same layout. Differentiating them depends largely on peculiar characters or dealing with components. This is cast as a problem of fine-grained classification.

While substantial research works can be found for text script identification in complex backgrounds [21-23], such methods are so far limited and have their own challenges. Pre-defined image classification algorithms, such as the robustly tested CNN [24] and the Single-Layer Networks (SLN) [25] normally consider holistic representation of images. Hence they perform poorly in distinguishing some script categories (e.g. English and Greek). The use of state of the art CNN classifiers for script identification is not straightforward, as they fail to counter the primary characteristic of extremely variable aspect ratio. Gomez et al [26] describe a new method using ensembles of conjoined networks as they form an important factor in a patch-based classification system. In [27], the authors proposed a novel approach, where Convolutional Neural Network (CNN) and Recurrent Neural Network (RNN) have been combined into an end-to-end framework.

Earlier, Attention mechanism has never been employed in script identification problem. However, in recent years, Attention model has been proved to be effective and impactful in the field of computer vision [59, 49, 60] and natural language processing [61]. But in the script identification task, few scripts are present which have similar character layouts. To distinguish them, attention in some specific areas is necessary. In this paper we introduce a novel feed forward attention mechanism for improving script identification. Attention improves the ability of the network to extract the most relevant information for each part of an input image. Thus it can also efficiently select those features which hold more significance at a particular step. To the best of our knowledge, ours is the first work to exploit attention mechanism for script identification task.

We used deep CNN architectures on image patches to extract their feature representations and eventually fed them to a LSTM network. After this, we used Attention mechanism for weight calculation of patches in order to give importance to those features which hold more significance. The patch-wise multiplication of these attention weights with the extracted CNN feature vectors yields the local features for individual patches whereas a global feature is obtained from the last cell state of LSTM. Local features contain fine-grained information while the global feature captures the holistic representation of the text images. Lastly we integrated local and global features using dynamic weighting because fusion of these features has been proved to give superior performance in various works [63]. In our work we employed attention based dynamic weighting to efficiently decide whom to assign more weightage between global feature and local feature, depending on their prominence. A fully connected layer is used at the end to obtain the classification scores for each patch. Final classification involves attention-wise summation of these patch-wise classification scores. Involving Attention at this step will allow the network to focus on relatively more important patches which would not have been possible if we used simple element-wise summation.

The major contributions of this paper are the following: (1) Both local and global features are extracted to preserve the fine-grained information as well as coarse-grained information of the images. (2) We propose a feed forward attention mechanism to assign weightage relatively between global and local features, according to their significance. Such a method allows the network to assign more importance to the least deformed parts of the image thus enabling the model to be more robust to noise. (3) Dynamic weighting of local and global features is used based on their contribution to the fused representation. Two different types of features together can effectively mitigate respective shortcomings of each feature. (4) Final classification involves attention-wise summation of patch-wise classification scores. It overcomes the limitation of element-wise summation which gives equal importance to all patches.

The rest of the paper is laid out as follows: In Section 2, we discuss some related works regarding development of script identification. In Section 3, the proposed attention based script identification framework has been described in details. In Section 4, we provide the experiment setup and discuss performance results in details. Finally, the conclusion is given in Section 5.

## 2. Related Work

Script identification is regarded as a well-described problem by document image analysis community. [28] provides a comprehensive review of various methods stated to tackle this problem. They classify the approaches into two main categories: techniques based on structure and visual appearance.

The techniques involving structure, require precise segmentation of text connected regions from the image, while methods relying on visual appearance are known for better performance in bi-level text. In the first category, Hochberg et al. [29] used cluster-based templates to handle unique characteristic shapes. Spitz and Ozaki [30,31] proposed to obtain the vertical distribution of concave outlines in connected components and then identify scripts at page-level, using their optical density. The authors of [32] considered both vertical and horizontal projection profiles and experimented on them for full-page document identification. Latest approaches in this division have obtained texture level features from Local Binary Patterns [36] or Gabor filters analysis [33-35]. Neural networks have been also employed [37,38] replacing hand-crafted features. All the methods mentioned above achieve high accuracy particularly for printed document images in mind. Also, some of them need large passages to extract sufficient information and hence do not perform well for scene text as they generally carry very less words.

Although extensive research has been done in script identification on printed document images, it is quite rare on non-conventional paper formats. Sharma et al. [39] relied on using conventional document analysing methods for identification of video-overlaid text at word stage. They study Gabor filters, Zernike moments, along with some hand-crafted gradient features earlier applied in tasks of handwritten character recognition. They overcame the in-built barriers of video-overlaid text by developing few dedicated algorithms for the pre-processing step. [21] deals with edge detection in overlaid-text images using a method that involves wavelet transform. After that some low-level features are extracted and they make use of a K-NN classifier. T.Q. Phan et al. designed algorithms and jointly performed both script identification and detection of video text overlay in [40]. They applied canny edge detection on text lines and evaluated connected components of those edges. Then they extracted upper and lower extreme points for each such component to analyze their texture properties like cursiveness. Shivakumara et al. [22,41] evaluated dominant gradients and explored their skeletons. They extracted a set of hand-crafted features after analysing the properties [41] of skeleton components, and studying the spatial [22] distribution of their branch points and end points. The above methods have been evaluated mainly for text appearing on video. Majority of these methods detect edges of text regions, which cannot be done for scene text. Also, Sharma et al. [42] identified video-overlaid text scripts at word level, by employing techniques based on Bag-of-Visual Words. They outperformed conventional script identification approaches that considered HoG as gradient based features or LBP as texture based features, by combining Bag-Of-Features (BoF) and Spatial Pyramid Matching (SPM) with patch based SIFT descriptors.

In 2015, the ICDAR Competition on Video Script Identification (CVSI-2015) [43] came up with a new standard dataset which tested the document analysis community. Apart from text images taken from videos of news, sports etc., it also included some examples of scene text. The most competitive pipelines in the contest were all built on CNN. They showed a considerable increase in accuracy as compared to methods based on hand-crafted features like HoG or LBP.

Shi et al. introduced Multi-stage Spatially-sensitive Pooling Network (MSPN) method in [44], where they provided the first real scene text images' dataset for script identification. The MSPN network's advantage is that unlike traditional CNNs, it does not require inputs to be of constant dimension. They achieved it by max pooling/average pooling along each row of the feature representations obtained at the intermediate levels. Their method is improved in [23] where they combined deep representations and mid-level features to design a globally trainable deep architecture. At every layer of the MSPN, local image descriptors were extracted with an encryption method that helped in CNN weight optimization. Nicolaou et al. [45] have presented a method based on hand-crafted features, a LBP variant, and a deep Multi-Layer Perceptron achieving superior performance in scene text script identification. Gomez et al. in [58] have proposed a patch-based method for script identification in scene text images. The method utilized patch-based CNN features, and the Naive–Bayes Nearest Neighbour classifier (NBNN). The same authors used a much deeper CNN framework in their extended work [26]. Moreover, they replaced the weighted NBNN classifier by a classification scheme based on patches. The new approach can be integrated in the CNN training procedure employing an Ensemble of Conjoined Networks. Thus their model had an advantage to learn simultaneously, both meaningful image patch feature maps and their individual significance in the patch-based classification rule. In [27], the authors trained together a CNN and a RNN into one globally trainable framework. CNN generates expressive feature maps, while RNN efficiently analyzes long-term spatial dependencies. Moreover, they handled input images of arbitrary sizes by adopting an average pooling structure. From all reviewed methods of script identification the one proposed here is the only one based on an attention-based patch weight classification framework. There are three key differences in the way we build our framework: (1) Use of Attention Network for weight calculation of patches to judge their priority according to information they contained, (2) Evaluation of both global and local features and (3) Dynamic weighting of local and global features, using fusion technique for successful script identification.

## 3. Proposed Framework

### 3.1. Overview

Provided a patch from an image I containing a few words, we estimate its script category $c \epsilon (1, \cdots, C)$. The brief overview of our proposed framework is illustrated in Figure 2. The end-to-end framework broadly contains three stages. In first stage, we use a stacked convolutional layers structure to extract precise translation invariant image features. The CNN layers generate varying dimension feature vectors. These vectors are fed into LSTM layer to utilize the spatial dependencies present in text script images. The second stage is an Attention network followed by softmax layer to obtain the patch weights. The reason for including Attention model is to give importance to those features which hold more significance. The patch-wise multiplication of this attention weights with the extracted CNN feature vectors yields the local features for individual patches. These local features contain fine-grained representation of the text images. To obtain the holistic information of these images, global feature is also extracted from the last cell state of the LSTM unit. Lastly we employed attention based dynamic weighting to integrate both local and global features, obtained in second stage. The classification scores for each patch are evaluated by using a fully connected layer at the end. Final classification involves attention-wise summation of these patch-wise classification scores to get final probability distribution over classes. It overcomes the limitation of element-wise summation which gives equal importance to all patches.

### 3.2. Review of CNN and LSTM Module

We first resize the height of the script image (containing few words) to a constant 40 pixels, maintaining the same aspect ratio. Then, we use sliding window approach to densely extract patches of size 32 × 32. The step size of the window is chosen as 8 pixels. For a particular image, starting from the left, we have extracted vertically two overlapping patches, thereafter shifted 8 pixels rightwards in the horizontal direction and carried out the same process successively. The particular values of the window scale and step size can be justified because they help in designing an improved scale invariant CNN architecture. The script length determines how many patches will be created. If D is the maximum patch count for a query image $X^{(i)}$ , then

$$X^{(i)} = (X_1^{(i)}, X_2^{(i)}, X_3^{(i)}, \dots, X_D^{(i)}) \tag{1}$$

where the superscript refers to $i^{th}$ sample and $X_d^{(i)} \epsilon \ \mathbb{R}^{32\times32}$ represents the individual patches.

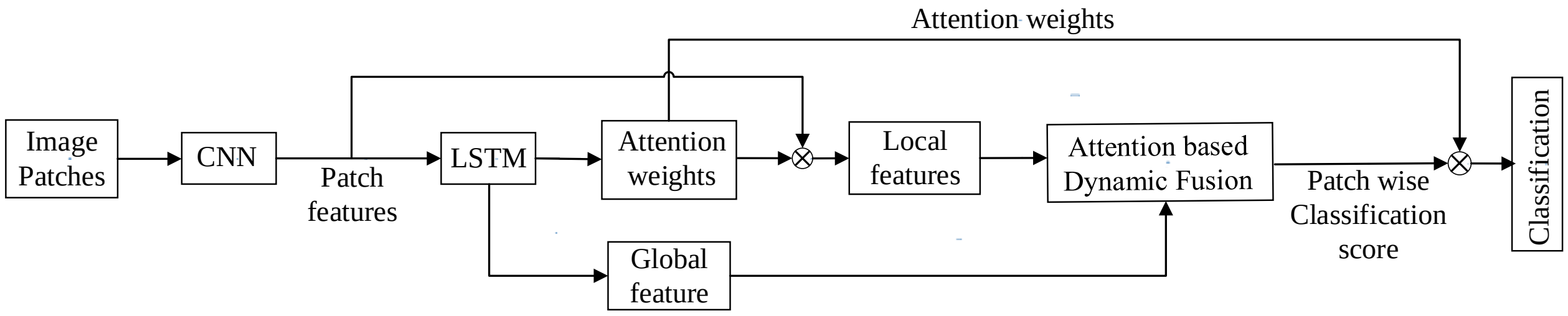

**Fig.2. Flowchart of our proposed framework**

Our framework design begins with the CNN architecture, which is used to obtain the text image representations. Each image patch is passed through this CNN network. The output response of the CNN for each patch in a given image $X^{(i)}$ is a 256 dimension feature vector.

$$Y_d^{(i)} = CNN\left(X_d^{(i)}\right) \forall \ d = 1 \to D \tag{2}$$

where $Y_d^{(i)} \epsilon \ \mathbb{R}^{256}$.

We used the CNN model proposed in [26] as it performed well in many instances for script identification. Our goal was to achieve the relevant CNN network that would provide optimum performance when integrated into our Attention model. Hence we varied the different parameters like number of convolutional layers, number of filters per layer, size of kernels and fully connected layers. Finally, we found that the CNN network in [26] gave the most promising results for script identification. The CNN model configuration has been provided in Table I. It contains three convolutional layers, each associated with pooling. Then an extra convolution layer is provided without pooling. Finally, the model ends with two fully connected layers.

The feature representations of all image patches, which are obtained after passing through a CNN network, are eventually fed to a LSTM following the same order of patch extraction. Spatial dependencies within text lines are overlooked by many of the previous approaches. However, it may be a critical step for script identification. RNN models are available which handle sequences and this allows the input text images to have arbitrary length. This naturally solves the issue while exploiting the spatial dependencies within text lines.

If an input vector x = $(x_1, x_2, \cdots, x_T)$ is provided, the RNN commonly used is:

$$h_t = f(x_t, h_{t-1}) \tag{3}$$

The hidden state $h_t$ simultaneously considers the current input $x_t$, as well as the earlier hidden state $h_{t-1}$ stored in the RNN block. The hidden states undergo a linear transformation to produce the RNN output.

Despite RNN being useful in dealing with sequence based problems, it has a disadvantage of vanishing gradient problem during back-propagation [46]. This restricts RNN's capability of handling considerably long contextual information. Vanishing gradient and exploding gradient are barriers in this task, due to presence of long text in script identification. The learning time increases and weights begin to oscillate, deteriorating the quality of the network. We redesign the unit using Long Short Term Memory (LSTM) [47] to elude the effect. LSTM addresses the issue by proposing three gating units: input, output and forget. These gates are incorporated into a block to model large long-temporal dependencies by preserving the gradient norm during back propagation. Input gate determines the amount of input information to be stored in hidden state. Output gate focuses on which hidden state information should be included in current time step output. Forget gate decides the hidden state information that should not be further remembered.

The gates operate based on the present input and previous hidden state. The hidden layer function is calculated using the following composite functions.

$$i_t = \sigma(\omega_{xi}x_t + \omega_{hi}h_{t-1} + \omega_{ci}c_{t-1} + B_i) \tag{4}$$

$$f_t = \sigma(\omega_{xf}x_t + \omega_{hf}h_{t-1} + \omega_{cf}c_{t-1} + B_f) \tag{5}$$

$$c_t = f_t c_{t-1} + i_t tanh(\omega_{xc}x_t + \omega_{hc}h_{t-1} + B_c) \tag{6}$$

$$o_t = \sigma(\omega_{xo}x_t + \omega_{ho}h_{t-1} + \omega_{co}c_t + B_o) \tag{7}$$

$$h_t = o_t tanh(c_t) \tag{8}$$

where i, o and f correspondingly denote the input, output and forget gates. σ refers to the logistic sigmoid function and c stands for cell. The subscripts of the weight matrix are self-explanatory like $\omega_{hf}$ which means the hidden-forget gate matrix, $\omega_{xo}$ means the input-output gate matrix etc. The cells to gate weight matrices are diagonal. A particular element A of the cell vector is the sole input to element A of each gate vector. The bias expressions $(B_i, B_f, B_c, B_o)$ have been discarded to keep simplicity.

Following the method in Shi et al. [23], we stacked two LSTM layers for better abstraction ability. The number of time steps in the LSTM layer depends on the number of patches obtained for each image. Hence time steps can vary from 1 to D. The output from each time steps is a 512 dimension feature vector.

$$h_d^{(i)} \epsilon \ \mathbb{R}^{512} \quad \forall\, d = 1 \to D \tag{9}$$

Back Propagation Through Time (BPTT) is chosen for learning of the parameters. Gradients are usually curtailed for clarity without disturbing the performance evidently. The overall process is shown in Figure 3. In the next section we will introduce an attention network to compute the patch weights.

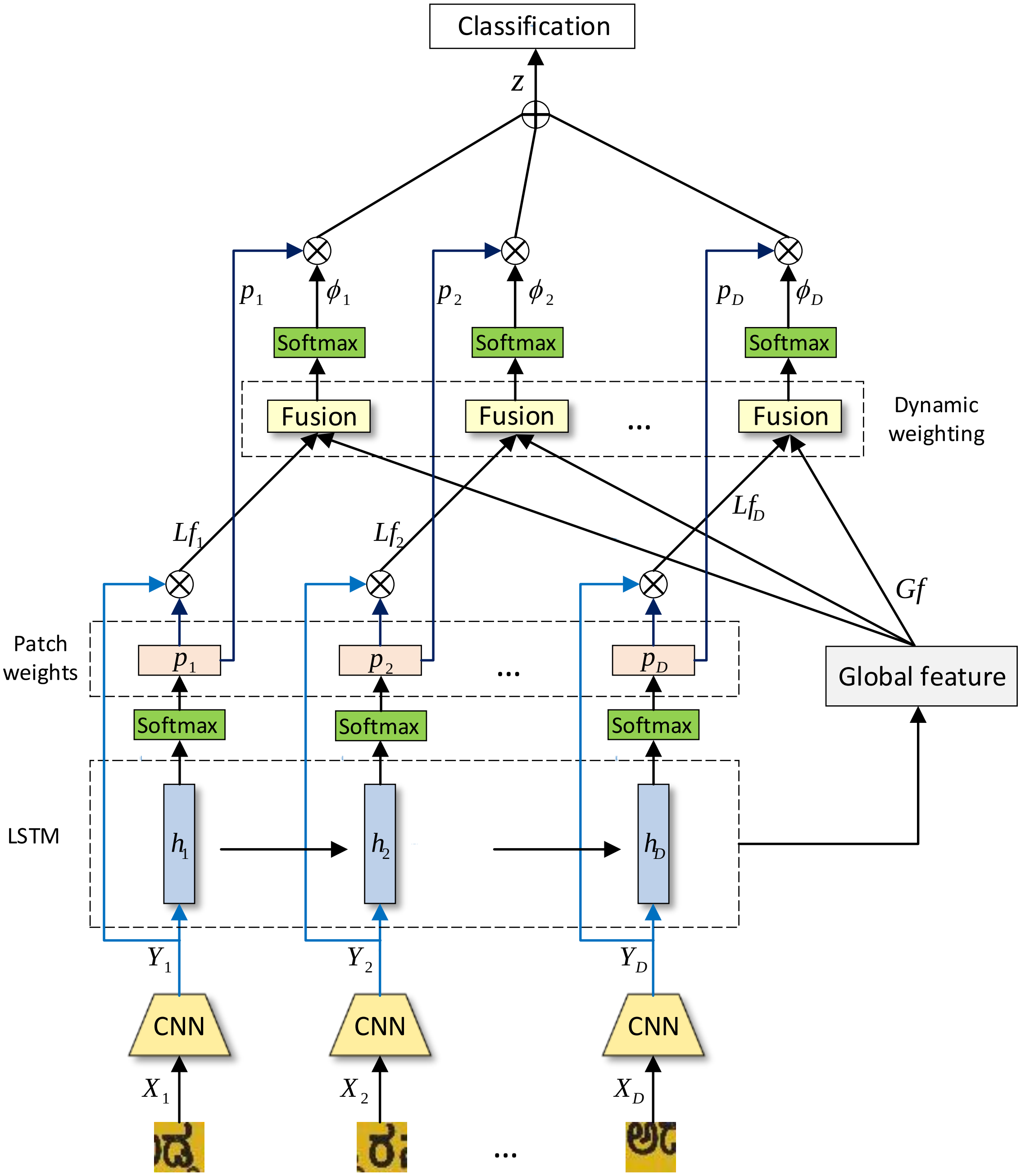

**Fig.3. Illustration of the training process of the proposed approach**

### 3.3. Attention based Patch weight calculation

Attention is a powerful mechanism that allows neural networks to focus on some particular portions of the input image in order to minimize the task complexity and discard irrelevant information. In the literature there are two types of attention [49]: "hard" attention and "soft" attention. In this work soft attention mechanism is employed. This means that we will be focusing everywhere at all times, but we will learn where to place more attention.

The output from the LSTM unit is passed through an attention network. The attention scores are computed by

$$q_d^{(i)} = v_a^T . \tanh( W_a . h_d^{(i)} + b_a ) \ \forall \ d = 1 \rightarrow D \tag{10}$$

where $W_a \epsilon \mathbb{R}^{256\times512}$ , $b_a \epsilon \mathbb{R}^{256}, v_a \epsilon \mathbb{R}^{256}$ are all trainable parameters.

Now these scores are tied to a softmax layer to produce the end probability weight distribution, such that the summation of all attention weights covering the required patches equals to 1.

$$p_d^{(i)} = \frac{\exp(q_d^{(i)})}{\sum_{d=1}^{D} \exp(q_d^{(i)})} \tag{11}$$

$$p_d^{(i)} = softmax\left(q_d^{(i)}\right) \ \forall d = 1 \rightarrow D$$

where [ $p_d^{(i)} \epsilon \mathbb{R}^D$ , $\sum_d p_d^{(i)} = 1$]

Thus, the attention weights are calculated for the patches. The patch-wise multiplication of these attention weights with the extracted CNN feature vectors yields the local features for individual patches. For certain scripts like English, Greek, Russian, some characters have similar layouts. Hence, it is necessary to capture some local patch specific information for discriminating them because global information is not sufficient in such scenarios. In other words, we intend to focus more on some of the specific patches which contain better script specific distinguishing features. The local features contain this fine-grained information of the text images. The reason for including Attention weights is to give importance to those features which hold more significance. For any image with D patches, local feature calculation:

$$Lf_d^{(i)} = p_d^{(i)} . Y_d^{(i)} \ \forall \ d = 1 \rightarrow D \tag{12}$$

To retain the holistic information of the images, a comprehensive feature representation is obtained from the last cell state of the LSTM unit. This is global feature of the entire sequence of patches. As stated in [47], LSTMs can be trained to link time intervals which are over 1000 steps even for noisy sequences without losing short-time-lag capabilities. Hence we can easily extract the global image representation from the last cell state which takes into account all the patches in a text line image. Though there are many local features for a particular image, there is only one global feature. Now that we have extracted both local and global features, each patch image is represented by two set of features $(Lf_d^{(i)}, Gf^{(i)})$. In the next section we deal with their dynamic fusion.

### 3.4. Dynamic weighting of Global and Local features

Local and global features are essential in representing an image. Local features generally hold the fine-grained information of objects, while global features represent the contextual information around objects. Thus integration of the local features and global features is an important step for script identification. These two features are combined to effectively improve the description accuracy. For fusing the two features at patch level, we introduce attention mechanism in our methodology. The low value of attention weight signifies less importance of that particular patch and subsequently we aim to prioritize the global feature using dynamic weighting in case of such instances. Similarly, the high value of attention weights encourages the network to give higher priority to local patch feature than the global holistic feature representation adaptively.

This module dynamically assigns weights to the two features $f_d^{(i)} \epsilon \left\{Lf_d^{(i)}, Gf^{(i)}\right\}$ by evaluating the coherence between them according to the following Equation:

$$\varphi_d^{(i)} = \sum_{k=1}^{2} c_{d,k}^{(i)} f_{d,k}^{(i)} \quad \forall\, d = 1 \rightarrow D \tag{13}$$

The coherence score $c_{d,k}^{(i)}$ are obtained in a similar way to the attention mechanism.

$$c_{d,k}^{(i)} = \frac{\exp(v_{d,k}^{(i)})}{\sum_{k=1}^{2} \exp(v_{d,k}^{(i)})} \tag{14}$$

where

$$v_{d,k}^{(i)} = w_k^T . \tanh\left(W_k . f_{d,k}^{(i)} + b_k\right)$$

(15)

where $\varphi_d^{(i)} \epsilon\ \mathbb{R}^{256}$ , $v_{d,k}^{(i)} \epsilon\ \mathbb{R}^1$ and $w_k^T, W_k, b_k$ are trainable parameters.

Through this manner the final feature representation of each patch image is obtained. The resulting feature maps of individual patches are then fed to a fully connected layer that has the same number of neurons as the number of classes. Finally, a softmax layer outputs the probability distribution over class labels for each patch.

$$\phi_d^{(i)} = softmax(W_f \varphi_d^{(i)} + b_f)\ \forall\ d = 1 \rightarrow D$$

(16)

where $\phi_d^{(i)} \epsilon\ \mathbb{R}^n$ , n is the number of class. Now the final decision rule would be weighted sum of $\phi_d^{(i)}$ over all the patches.

$$z^{(i)} = \sum_{d=1}^{D} p_d^{(i)} . \phi_d^{(i)}$$

(17)

where $p_d^{(i)}$ is the attention weight we obtain earlier. In this way, we obtain the final probability distribution $z^{(i)}$ over all the classes for a query image. The following average negative log-likelihood error over the training set combined with a regularization term yields the cost function.

$$L\big(Z^{(i)}, w\big) = \frac{1}{N} \sum_{i=1}^{N} [-\, Z^{(i)} \log(z^{(i)})] + \lambda \|w\|_2^2$$

(18)

where, $Z^{(i)}$ is the ground truth of the word image, $w$ represents the learning weights, $\lambda$ is weight decay parameter and $N$ is the number of word images in a particular batch.

Please note that the proposed framework is an end-to-end network where the model takes the image patches extracted from an input text line/word image as input, and at the end it gives the final class distribution of that particular text line/word image. We impose the supervision with respect to every text line/word image using the loss function mentioned in eqn. 18 in order to train the network in an end-to-end manner. We follow a particular patch extraction strategy where a 32 × 32 window slides over the entire image with a stride of 8 in the both vertical and horizontal direction. Not all the patches are equally important for discriminating a particular script and therefore attention mechanism helps to calculate the relative importance of the image patches by assigning a weight to all the patches. Hence, a CNN network is used to extract a 256 dimensional latent feature vector from each image patch. Thereafter, these feature representations are fed to LSTM following the same order of patch extraction in order to obtain the attention weights. We use the attention weights for two times – (1) at first it is multiplied with the patch features to obtain the local level features. Low value of attention weight will cause the local features to be less important for that particular patch and adaptively give more priority to global feature through dynamic weighting. (2) The same weights are also used in the last step while computing the final classification results. This will force the network to learn the relative importance of the image patches and overcome the limitations of using simple element-wise summation.

**Algorithm1.** Script identification in natural scene text images and video scripts

---

**Input:** Natural scene text images converted to **D** patches of size 32x32
**Output:** Identified script.

**For** each patch $\mathbf{X_d}$ of $X_1...X_D$ **do**

Step 1: Feed patch $\mathbf{X_d}$ into CNN and obtain feature vector $\mathbf{Y_d}$ as output

Step 2: $\mathbf{Y_d}$ is given as input to the $\mathbf{d^{th}}$ cell state of LSTM

Step 3: Obtain attention weight $\mathbf{p_d}$ as output from LSTM

Step 4: Patch wise multiplication of $\mathbf{p_d}$ with $\mathbf{Y_d}$ to extract local feature $\mathbf{Lf_d}$

$$Lf_d = p_d . Y_d$$

Step 5: If d = D, extract global feature **Gf** as output from the last cell state of LSTM

**End for**

**For** each patch $\mathbf{X_d}$ of $X_1...X_D$ **do**

Step 6: Dynamic weighting of global feature **Gf** with local feature $\mathbf{Lf_d}$ and apply fully connected layer to classify into scripts

**End for**

Step 7: Final classification which involves attention based weighted summation of **D** classifications $z = \sum_{d=1}^{D} p_d . \phi_d$

---

## 4. Experiments

### 4.1 Datasets

There exist many datasets [50,40,51] containing scripts of different languages. In this work we evaluated our proposed model over four multilingual video word datasets – CVSI-2015, SIW-13, ICDAR-2017 and MLe2e dataset. The CVSI-2015 [43] dataset contains scene text images of ten different scripts: English, Hindi, Bengali, Oriya, Gujrati, Punjabi, Kannada, Tamil, Telegu, and Arabic. Each script has at least 1,000 text images collected from different sources (i.e. news, sports etc.). The dataset has three sets – training set (60%), validation set (10%) and test set (30%).

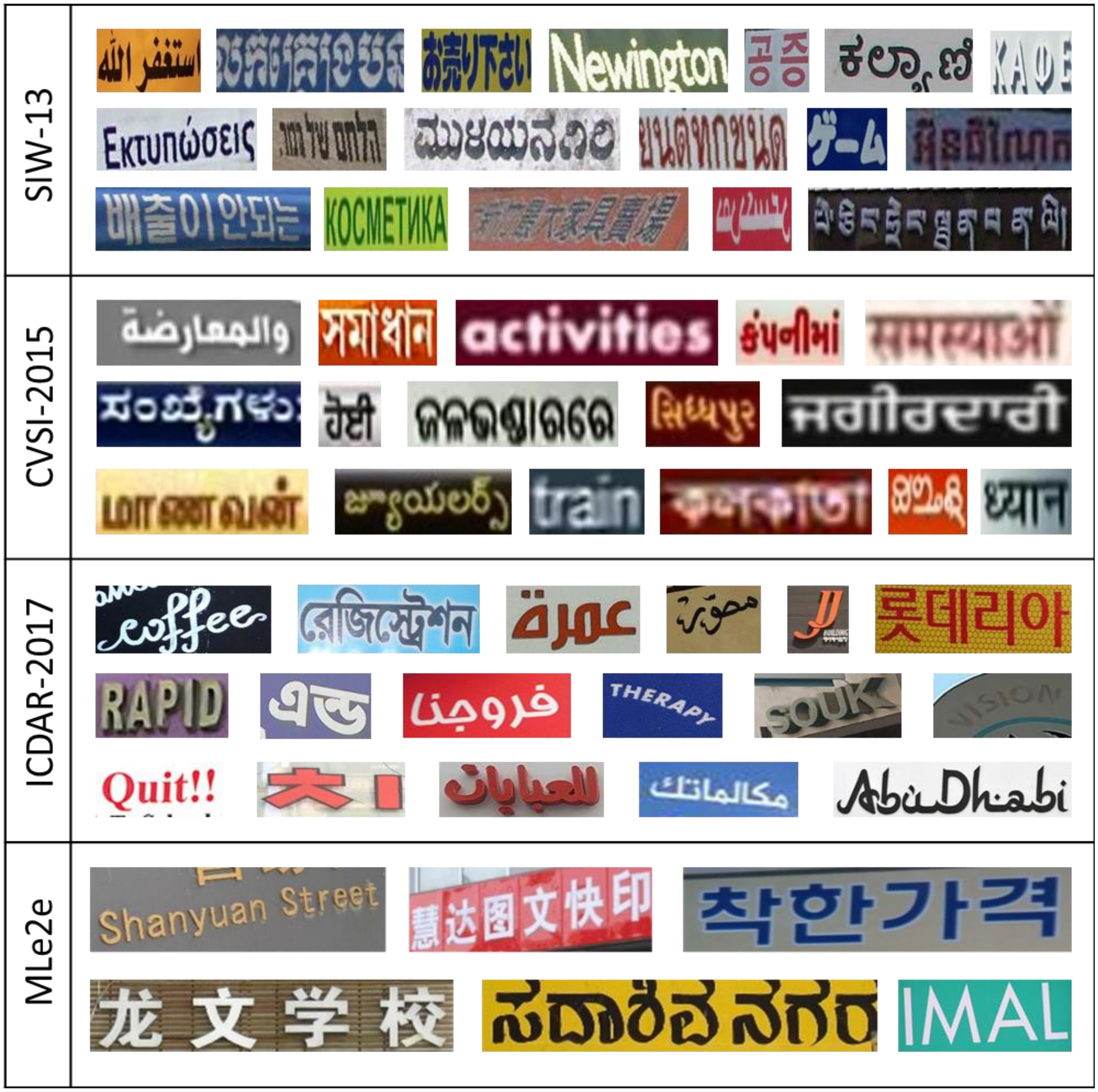

**Fig.4. Examples of scene text scripts in the SIW-13, CVSI-2015, ICDAR2017 and MLe2e datasets**

The SIW-13 dataset [23] consists of 16,291 multi-scripts text images in 13 classes: Arabic, Cambodian, Chinese, English, Greek, Hebrew, Japanese, Kannada, Korean, Mongolian, Russian, Thai, and Tibetan. The images are collected from Google street view. Some samples of this dataset are shown in Figure 4. Since they are natural scene images, the texts appearing in the images are in different orientation, fonts, colour and size. These factors make the datasets much more challenging for script identification task.

The ICDAR-2017 [64] dataset has 68,613 cut out word images for training. The validation set has 16,255 word images. The dataset consists of 9 languages Arabic, English, French, Chinese, German, Korean, Japanese, Italian, Bangla. Out of the above languages English, French, German, Italian share the same Latin script. However, in our current work, these scripts are assigned the same script class: Latin. Additionally, isolated punctuation or other special characters are considered as a special script class, namely Symbols. Hence, we have total 7 script classes.

We also used the MLe2e [26] dataset which is considered as a Multi-Language end-to-end dataset for the evolution of the scene text images starting from text region detection to script identification and text recognition tasks. But, as we are more interested in script identification task, we used the pre-segmented text version of the dataset containing the cropped word images. The dataset contains 1178 and 643 word images for training and testing respectively of four different scripts, namely Latin, Chinese, Kannada, and Hangul. Some examples are shown in Figure 4.

### 4.2. Implementation Details

Here, we describe the architecture of the model used in this paper. To achieve the optimum CNN architecture that would fit into our model, we varied necessary parameters and tested the different versions of CNN on CVSI-2015 dataset. The following parameters were tuned in this procedure: the size and step of the sliding window, the base learning rate, the number of convolutional, the number of neurons in the fully connected layers, the convolutional kernel sizes, and the feature map normalisation schemes. Finally, we concluded that the CNN architecture proposed in [26] gave the most promising results for our end-to-end model. The configuration of the CNN model is summarized in Table I.

**Table I. Network configuration of the basic CNN model**

| Type` | Configuration |
|---|---|
| Input | 32 × 32 patches |
| Convolution | Filters: 96, kernel size: 5 × 5, Stride: 1,<br>Output size: 96 × 28 × 28. |
| Max pooling | kernel size: 3, stride: 2, pad: 1,<br>Output size: 96 ×15 × 15. |
| Convolution | Filters: 256, kernel size: 3 × 3, Stride: 1,<br>Output size: 256 × 13 × 13. |
| Max pooling | kernel size: 3, stride: 2, pad: 1,<br>Output size: 256 ×7 × 7. |
| Convolution | Filters: 384, kernel size: 3 × 3, Stride: 1,<br>Output size: 384 × 5 × 5. |
| Max pooling | kernel size: 3, stride: 2, pad: 1,<br>Output size: 384 ×3 × 3. |
| Convolution | Filters: 512, kernel size: 1 × 1, Stride: 1,<br>Output size: 512 × 3 × 3. |
| Fully connected layer | 4096 neurons |
| Fully connected layer | 256 neurons |

After CNN, we implemented a simple 2-layer LSTM model. Then a Softmax layer is used to obtain attention patch weights. This part has the following configuration:

- layer1: 512 hidden LSTM units
- layer2: 512 hidden LSTM units
- Softmax layer

To prevent over-fitting dataset is enlarged using data augmentation. We use same CNN parameters for all the patches. We initialize the weights of the model according to the Xavier initializer [52]. Rectified Linear Units (ReLU) [53] is applied after the convolution and fully connected layers. Batch normalization [54] is employed to effectively increase the training speed. The architecture associates the dropout [55] strategy with fully connected layers. The dropout rate was maintained at 0.5 throughout training. The network is built with ≈ 12M parameters. However, usage of deeper and wider convolutional layers can be beneficial in extracting more complicated features.

We implement our framework in TensorFlow on a server with Nvidia Titan X GPU. Optimization of the network is done with Adam Optimizer. The model is trained for 20k iterations with batch size 32 and learning rate 0.001. The weight decay regularization parameter is fixed to $5 \times 10^{-4}$. The computational cost increases with the length of the images resulting more time to converge. Usually the number of patches for each image varies from 10 to 60 in the CVSI dataset. But for a lengthy script this number goes beyond 100. Thus maximum number of patches allowed is set to a threshold value N. If the number of patches is more than this threshold, then we will choose randomly N patches. In our experiments we take value of $N$ as 100. Also batch size was reduced to 32 to accommodate the GPU's memory. During evaluation we noticed that each image takes roughly 85ms on average on GeForce Titan X.

### 4.3. Baseline approaches

We compare our proposed method with several baseline methods including some traditional approaches like LBP, Basic CNN, Single-Layer Network, MSPN, DisCNN, Convolutional Recurrent Neural Network and Ensembles of Conjoined Networks.

(1) Local Binary Patterns (LBP): LBP [56] is a widely adopted texture analysis technique. Fixed face images are divided into several 8 x 8 grids. LBP features are extracted from them using the vl_lbp function in the VLFeat library [57]. These features when combined into a new 2784-dimension vector act as image descriptor. Finally, they classify using simple SVM.

(2) Basic CNN (CNN): A traditional CNN architecture, named CNN-Basic, is also used as a baseline. Since the fully connected layers are present, only fixed dimension images (here samples are cropped to 100 x 32) can be fed into a conventional CNN structure. SGD is adopted for training the CNN-Basic.

(3) Single-Layer Networks (SLN): In [25] Coates et al. proposed a simple unsupervised feature learning technique using K-Means clustering to obtain state-of-the-art results in image classification. We extracted features using the feature learning code made public by the authors.

(4) MSPN: Multi-Stage Pooling Network as proposed in [44]. It has architecture of CNN network that contains multiple stage horizontal pooling. The outputs of the three pooling layers are concatenated as a long vector, which is fed to later fully-connected layers. We use the same architecture as used in [44] for comparisons.

(5) DisCNN: In [48], deep representations and mid-level features are jointly trained into an end-to-end deep network. Training the images with a pre-defined CNN architecture, we densely extract the local deep feature maps. Based on the learned discriminative patterns, mid-level representation is derived by encrypting the local features.

(6) Convolutional Recurrent Neural Network (CRNN): In [27] they combined a Convolutional Neural Network (CNN) and a Recurrent Neural Network (RNN) into a globally trainable deep model. The CNN network generates expressive image representations, while the RNN module helps to efficiently handle input images of arbitrary sizes.

(7) Ensembles of Conjoined Networks (ECN): In [26] a patch based classification method is introduced. Image patches are obtained from the input images following a certain sampling strategy. Feature representation of each patch is obtained by using a deep CNN architecture. They used a simple global decision rule that takes average of the output feature representation of the network for all patches in a given script image.

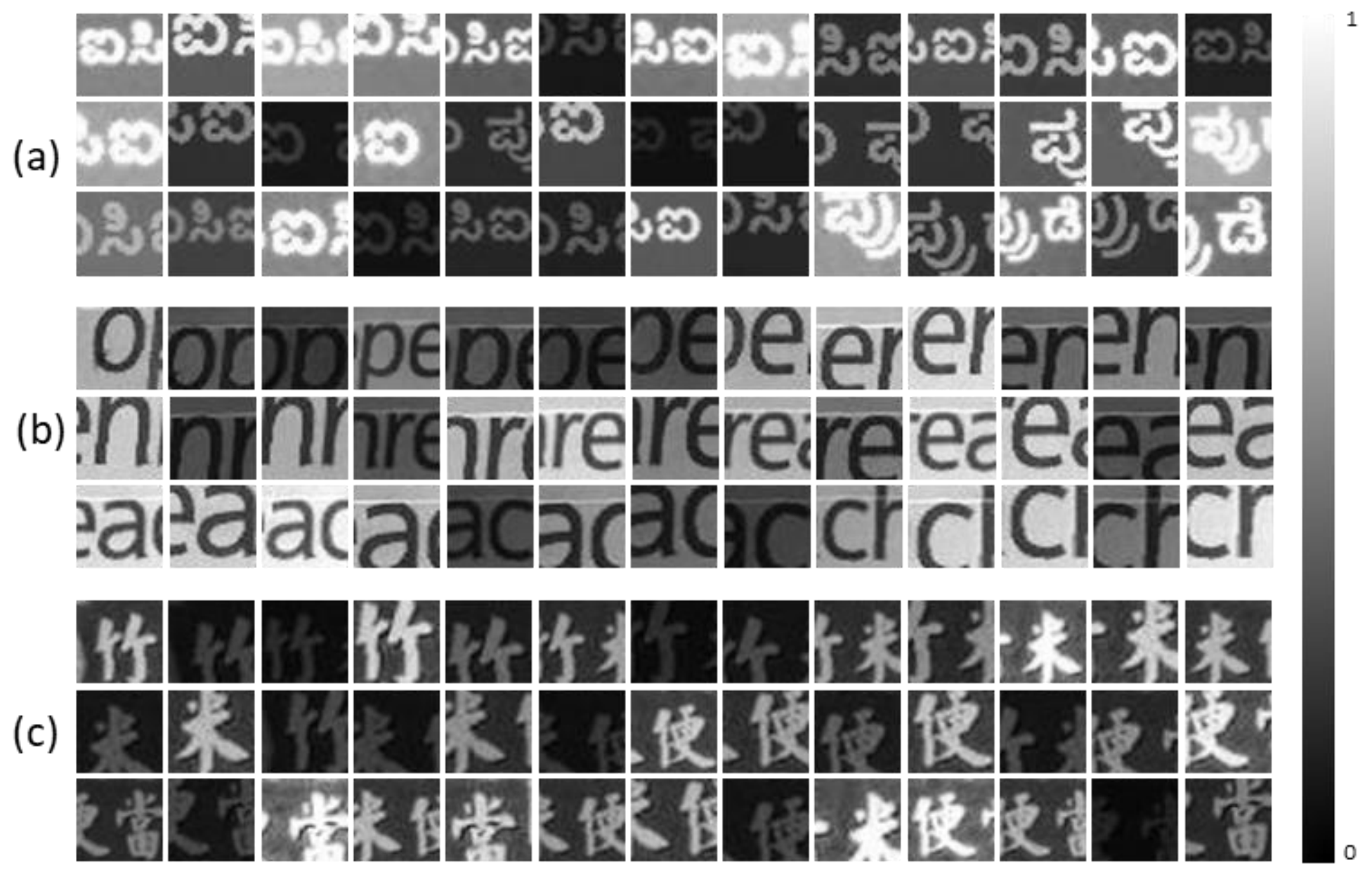

**Fig.5. The generated Attention maps for three scripts from SIW-13 dataset. – (a) Kannada (b) English and (c) Chinese**

### 4.4. Experiments in SIW-13 dataset

We train and test our model on the SIW-13 dataset consisting of 13 scripts. For comparison, we evaluate seven other methods as described in the baseline section.

The results in this dataset using our method and the baseline methods are illustrated in Table II. From Table II we can see that, the proposed method consistently outperforms other methods. The LBP approach performs well on those scripts which have larger appearance differences. They are easier to distinguish via texture features. The method does not perform well on scripts containing certain characters that have strikingly similar layout. LBP is bettered by both Basic-CNN and SLN in logographic type scripts. But on the similar-subset scripts, they also do not perform that well. Then we evaluated DisCNN and found that it leads to improvement over previous methods in almost all types of scripts. More recent methods CRNN and ECNN which employed CNN-RNN fused deep networks and patch based CNN network respectively, not only bettered performance in logographic scripts but also brought huge change in the Alphabetic scripts like English, Greek, Russian etc. Comparing the accuracies of different script classes, we have shown the confusion matrix in figure 8.

Here it is noticed that Arabic scripts and Thai scripts have higher accuracies than that on other languages. The uniqueness in writing styles is the reason why these scripts can be easily differentiated from other scripts. However, scripts like Greek, English, Russian etc. that are mostly based on Latin, are comparatively more challenging for identification. On these scripts, all previous methods obtain lower accuracies because they have similar holistic representation. This makes it more challenging to identify these scripts. But, in our method attention allows the network to focus on more relevant and discriminative part of the scripts. Our framework slightly improved the performance of these methods. In figure 5 some samples of generated attention maps are shown. We can visualize the relative importance of the patches from the attention maps. High attention is shown in white and low attention is shown in black.

**Table II. Script wise results of different methods on SIW-13**

| Script | LBP | CNN | SLN | MSPN | DisCNN | CRNN | ECN | Our method |
|---|---|---|---|---|---|---|---|---|
| **Ara** | 64.0 | 90.0 | 87.0 | - | 94.0 | 96.0 | 98.0 | **99.0** |
| **Cam** | 46.0 | 83.0 | 76.0 | - | 88.0 | 93.0 | 99.0 | **99.0** |
| **Chi** | 66.0 | 85.0 | 87.0 | - | 88.0 | **94.0** | 88.0 | 92.0 |
| **Eng** | 31.0 | 58.0 | 64.0 | - | 71.0 | 83.0 | 97.0 | **98.0** |
| **Gre** | 57.0 | 70.0 | 75.0 | - | 81.0 | 89.0 | 99.0 | **100.0** |
| **Heb** | 61.0 | 89.0 | 91.0 | - | 91.0 | 93.0 | 97.0 | **99.0** |
| **Jap** | 58.0 | 75.0 | 88.0 | - | 90.0 | 91.0 | 92.0 | **98.0** |
| **Kan** | 56.0 | 82.0 | 88.0 | - | 91.0 | 91.0 | 89.0 | **92.0** |
| **Kor** | 69.0 | 90.0 | 93.0 | - | **95.0** | **95.0** | 90.0 | 93.0 |
| **Mon** | 77.0 | 96.0 | 95.0 | - | 96.0 | 97.0 | 94.0 | **98.0** |
| **Rus** | 44.0 | 66.0 | 70.0 | - | 79.0 | 87.0 | **95.0** | 93.0 |
| **Tha** | 61.0 | 79.0 | 91.0 | - | 94.0 | 93.0 | **95.0** | **95.0** |
| **Tib** | 88.0 | 97.0 | 97.0 | - | 97.0 | **98.0** | 97.0 | 97.0 |
| **Average** | 60.0 | 82.0 | 85.0 | 86.0 | 89.0 | 92.0 | 94.0 | **96.5** |

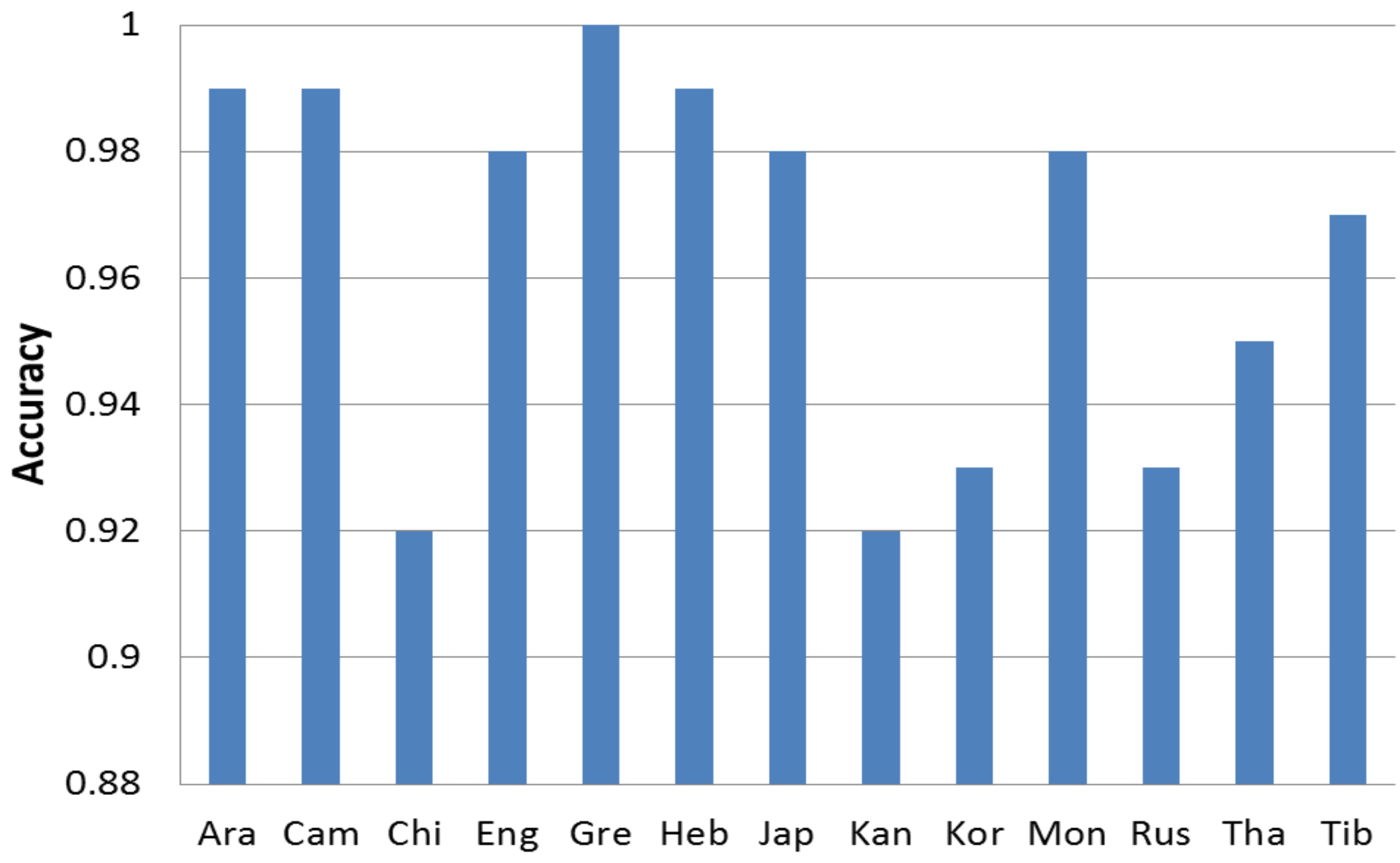

**Fig.6. Graphical representation of script wise performance on SIW-13**

### 4.5. Experiments in CVSI-2015 dataset

We also tested our method on CVSI-15 dataset. Of all text images, 60% assigned for training, 10% for validation and the remaining 30% are for testing. CVSI2015 is relatively more simple, with limited variation compared with SIW13 dataset. Table III compares our method with the baseline methods on CVSI2015. As noticed, Google performs the best while our method also achieves competitive accuracy for the task. Google's demerit is that it applies image pre-processing method based on binarization. This works fine for only text that has great background, limiting the method's ability for text identification in natural scenes. However, our method does not suffer from this drawback. It can be used for complex background and also for slightly distorted images. HUST [44] also achieves a high accuracy due to usage of multiple features. Our model applies local and global features with further dynamic weighting on them. Thus, our model is able to achieve better performance. ECN [26] is also able to obtain a good result. But the main drawback of this method is that they treat all the image patches equally, irrespective of whether they contain relevant information or not. Our method uses attention mechanism to give relative importance to the patches. Thus our network is able to achieve a better classification result.

Table III. Script wise results of different methods on CVSI-15

| Script | Google | C-DAC | HUST | CVC-2 | CUK | ECN | Our method |
|---|---|---|---|---|---|---|---|
| **Eng** | **97.95** | 68.33 | 93.55 | 88.86 | 65.69 | - | 94.20 |
| **Hin** | **99.08** | 71.47 | 96.31 | 96.01 | 61.66 | - | 96.50 |
| **Ben** | **99.35** | 91.61 | 95.81 | 92.58 | 68.71 | - | 95.60 |
| **Ori** | **98.47** | 88.04 | **98.47** | 98.16 | 79.14 | - | 98.30 |
| **Guj** | 98.17 | 88.99 | 97.55 | 98.17 | 73.39 | - | **98.70** |
| **Pun** | **99.38** | 90.51 | 97.15 | 96.52 | 92.09 | - | 99.10 |
| **Kan** | 97.77 | 68.47 | 92.68 | 97.13 | 71.66 | - | **98.60** |
| **Tam** | 99.38 | 91.90 | 97.82 | **99.69** | 82.55 | - | 99.20 |
| **Tel** | **99.69** | 91.33 | 97.83 | 93.80 | 57.89 | - | 97.70 |
| **Ara** | **100.00** | 97.69 | **100.00** | 99.67 | 89.44 | - | 99.60 |
| **Average** | **98.91** | 84.66 | 96.69 | 96.00 | 74.06 | 97.2 | 97.75 |

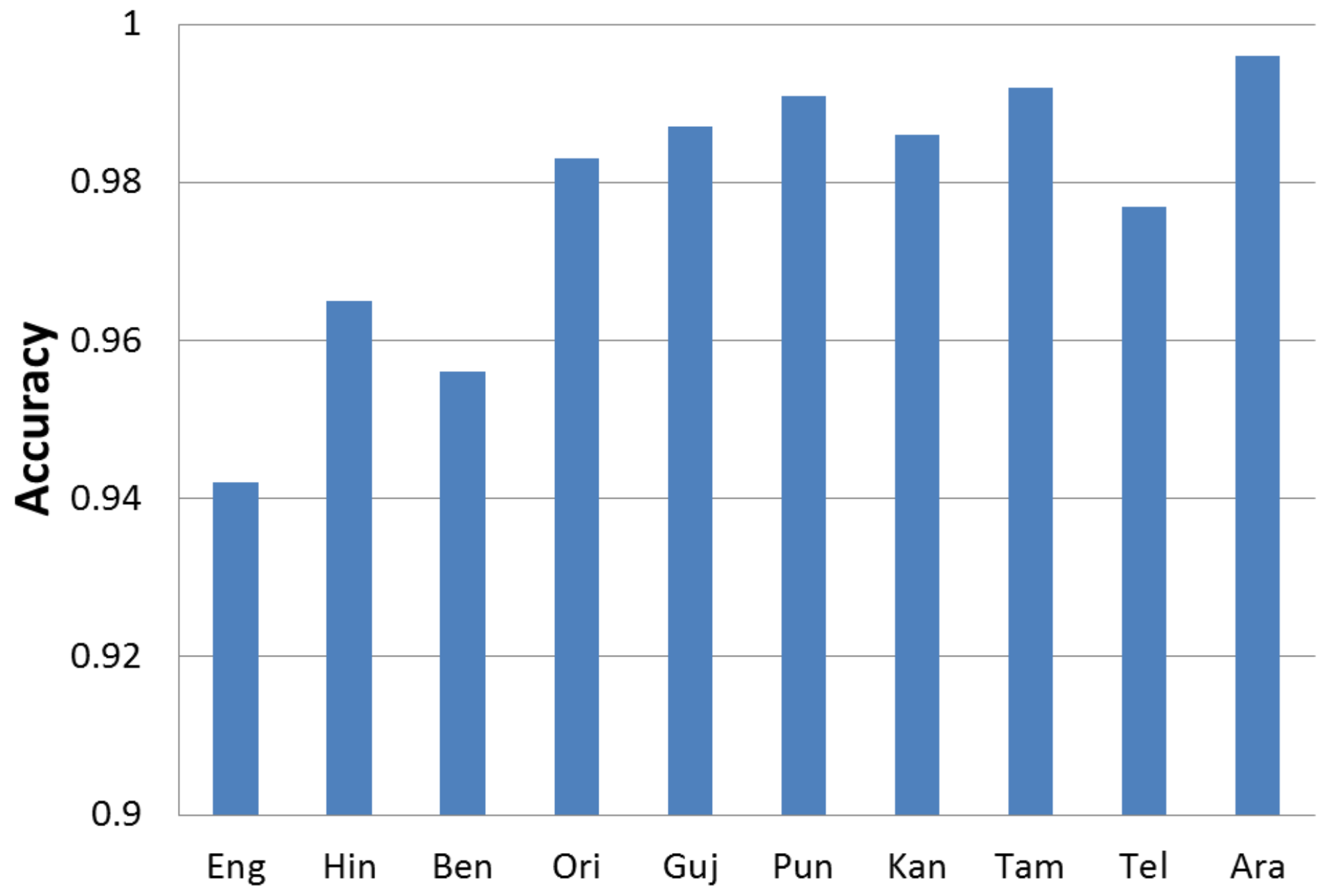

**Fig.7. Graphical representation of script wise performance on CVSI-15**

### 4.6. Experiments in ICDAR-2017 dataset

We also evaluated our method on ICDAR-2017 dataset. It consists of 7 script classes. The dataset is quite large as compared to other datasets. The validation set is used for evaluating our model. We compare the results with several other methods like E2E-MLT [65] and ECN [26]. We have implemented those methods and obtained the script identification accuracies on validation dataset as reported in Table IV. E2E-MLT uses a VGG-16 model pre-trained on ImageNet dataset along with Global Average Pooling layer after the final convolution layer. This method performs moderately on the dataset. Accuracy obtained from this method is below 90%. It is evident that this dataset is more challenging with complex background and stylish fonts. ECN [26] also could not perform well on this dataset. The proposed method outperforms all the previous entries and increases the classification accuracy on this dataset by 2%. Hence, the result justifies that the proposed additional complexity is worth.

**Table IV. Script identification results of different methods on ICDAR-17 dataset**

| **Method** | **Accuracy (%)** |
|---|---|
| **ECN** [26] | 86.46 |
| **E2E-MLT** [65] | 88.50 |
| **Our method** | **90.23** |

### 4.7. Experiments in MLe2e dataset

The results on MLe2e dataset is illustrated in Table V. It shows that our method performs well on the dataset. It is noticed that, with increasing complexity of the script datasets, the use of attention mechanism and fusion of local and global features to obtain a robust feature representation become more important for better performance. The confusion matrices of all the four datasets are shown in Figure 8.

**Table V. Script identification results of different methods on MLe2e**

| Method | Accuracy (%) |
|---|---|
| **CVC-2** [43] | 88.16 |
| **Gomez** [58] | 91.12 |
| **ECN** [26] | 94.40 |
| **Our method** | **96.70** |

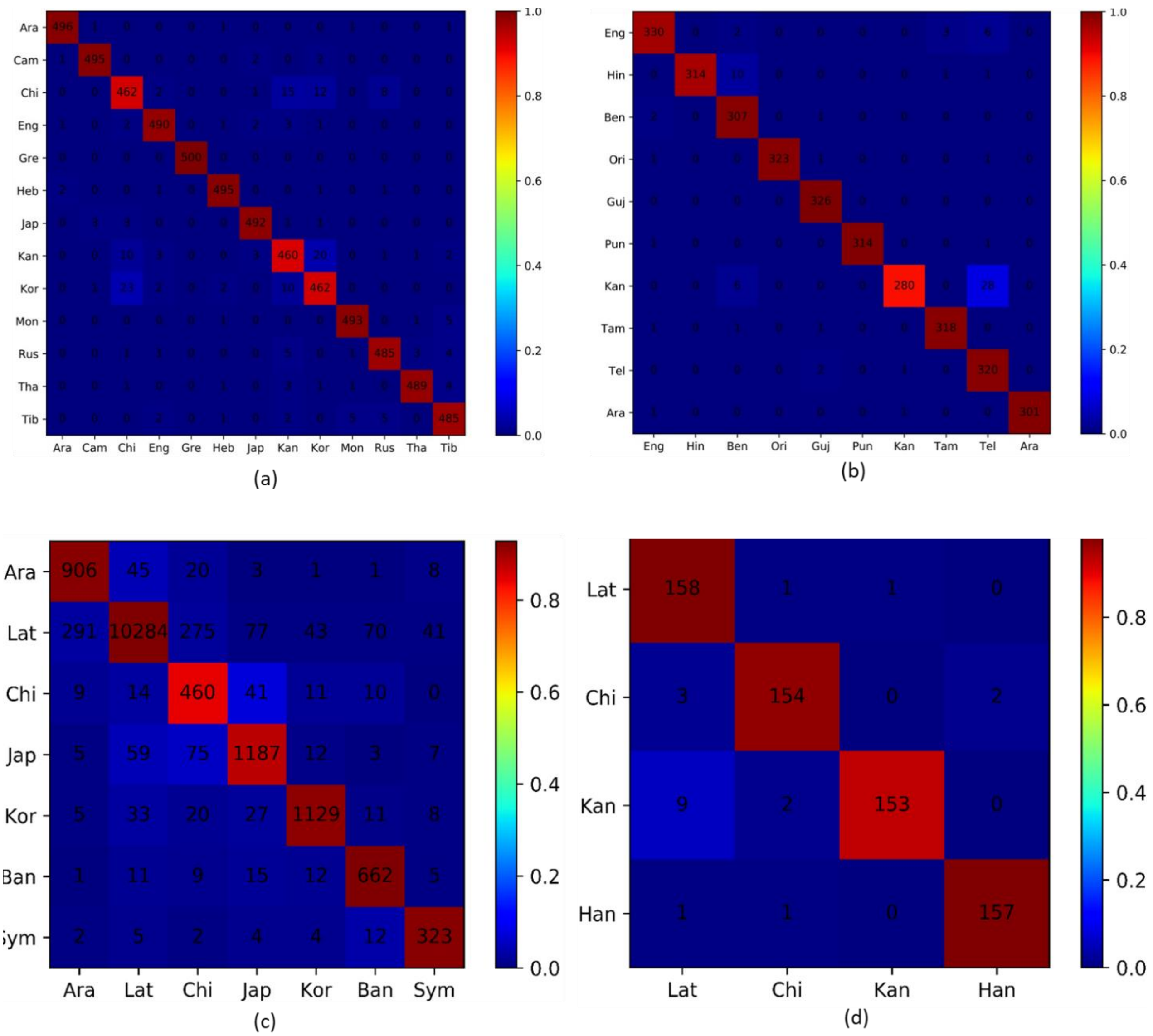

**Fig.8. Confusion matrices for four datasets (a) SIW-13 (b) CVSI-2015 (c) ICDAR-2017 (d) MLe2e**

### 4.8. Improvement Analysis

In this section, we provide breakdown of different sub-variants of our configuration and tested them to analyze the gradual improvement. The results are summarised in Table VI.

**#Variant-1:** This variant of our configuration was designed with only a CNN-LSTM framework without any attention model. This method extracted local features from patches without assigning any attention weights. Thereafter, they were simply concatenated with global features and classified into required classes. It did not perform so well. This was because due to absence of attention mechanism it treated all patches equally, irrespective of whether they provided more details or fewer details.

**#Variant-2:** This variant used the attention only once and it was in CNN-LSTM framework. The patch-wise multiplication of the attention weights with the extracted CNN feature vectors yields the local features for individual patches. Global features were also evaluated. This was followed by simple concatenation of global and local features and classification into required classes. On applying attention during generation of local features in CNN-LSTM, we could focus more on those patches which hold more significance. Due to this change, script identification results improved slightly in images having complex background and distortion issues. On the other hand, simple fusion was a bad approach since feature vectors became high-dimensional. Redundancy crept in, leading to longer processing times. For an image patch we were unable to take into consideration, whether its local feature should be given more priority or the global feature.

In this way, we zeroed into our final architecture, which outperformed the previous variants by a large margin. Our architecture had attention in CNN-LSTM design for generation of local features, which helped while performing dynamic weighting of local and global features. Unlike previous variants, during fusion we could decide for a patch, the relative importance of its local and global features. Again Attention was involved in summation of patch-wise classification scores overcoming the element-wise summation approach that treated all patches equally.

**Table VI. Results of different variants of our configuration on different datasets**

| Dataset | Variant | Configuration | Accuracy (%) |
|---|---|---|---|
| **SIW-13** | **Variant 1** | CNN+LSTM(no attention)+Fusion(concatenation) | 94.10 |
| | **Variant 2** | CNN+LSTM(with attention)+ Fusion(concatenation) | 95.90 |
| | **Our method** | CNN+LSTM(with attention)+Dynamic weighting | **96.50** |
| **CVSI-15** | **Variant 1** | CNN+LSTM(no attention)+Fusion(concatenation) | 97.25 |
| | **Variant 2** | CNN+LSTM(with attention)+ Fusion(concatenation) | 97.65 |
| | **Our method** | CNN+LSTM(with attention)+Dynamic weighting | **97.75** |
| **ICDAR-17** | **Variant 1** | CNN+LSTM(no attention)+Fusion(concatenation) | 87.30 |
| | **Variant 2** | CNN+LSTM(with attention)+ Fusion(concatenation) | 89.14 |
| | **Our method** | CNN+LSTM(with attention)+Dynamic weighting | **90.23** |
| **MLe2e** | **Variant 1** | CNN+LSTM(no attention)+Fusion(concatenation) | 94.36 |
| | **Variant 2** | CNN+LSTM(with attention)+ Fusion(concatenation) | 95.13 |
| | **Our method** | CNN+LSTM(with attention)+Dynamic weighting | **96.70** |

## 5. Conclusion

In this paper, we presented a novel method for script identification in natural scene text images and video scripts. We are the first to introduce Attention mechanism in script identification. The method generates local features through attention-based patch weighting scheme and thereby performs dynamic weighting of local and global features using dynamic weighting technique. It is fed to a fully connected layer to get classification scores. Finally, an attention-wise summation is carried out on all patch-wise classification scores. Experiments performed in four datasets demonstrate state of the art accuracy rates in comparison with other recent approaches. It is worth mentioning that our algorithm handles many common drawbacks very well. It achieves better performance when dealing with complex background, distortion, low resolution of images.

During our experiments, we have noticed that ICDAR 2017 script identification dataset contains four different language scripts, namely, English, French, Italian, and German. However, all these four different language scripts have been labelled only as Latin script. The exact language script is extremely important in order to recognize the word image, since most of the state-of-the-art word recognition models are language dependent. Henceforth, in scenarios where a common script is used by multiple languages, identification of the exact language would be an interesting future research direction. In future we are also looking forward to implement an end-to-end method which jointly tackles the problem of multi-lingual text detection and script identification to make the system more robust.